\documentclass[preprint,authoryear,12pt]{elsarticle}
\usepackage{mathrsfs}
 \usepackage{graphics,graphicx,epsfig}

\usepackage{amsmath,mathrsfs,amssymb,amsfonts}
\usepackage{algorithm,algorithmic}
\usepackage{multirow}
\newtheorem{theorem}{Theorem}
\newtheorem{lemma}{Lemma}

\newtheorem{corollary}{Corollary}

\newenvironment{pf}{\noindent{\it Proof: }}{\qed\medskip}

\journal{Artificial Intelligence Journal}

\begin{document}

\begin{frontmatter}

\title{On the Doubt about Margin Explanation of Boosting}

\author{Wei Gao and Zhi-Hua Zhou\corref{cor1}}

\address{National Key Laboratory for Novel Software Technology\\
             Nanjing University, Nanjing 210023, China}
\cortext[cor1]{Email: zhouzh@lamda.nju.edu.cn}
\begin{abstract}
Margin theory provides one of the most popular explanations to the success of \texttt{AdaBoost}, where the central point lies in the recognition that \textit{margin} is the key for characterizing the performance of \texttt{AdaBoost}. This theory has been very influential, e.g., it has been used to argue that \texttt{AdaBoost} usually does not overfit since it tends to enlarge the margin even after the training error reaches zero. Previously the \textit{minimum margin bound} was established for \texttt{AdaBoost}, however, \cite{Breiman1999} pointed out that maximizing the minimum margin does not necessarily lead to a better generalization. Later, \cite{Reyzin:Schapire2006} emphasized that the margin distribution rather than minimum margin is crucial to the performance of \texttt{AdaBoost}. In this paper, we first present the \textit{$k$th margin bound} and further study on its relationship to previous work such as the minimum margin bound and Emargin bound. Then, we improve the previous empirical Bernstein bounds \citep{Maurer:Pontil2009,Audibert:Munos:Szepesvari2009}, and based on such findings, we defend the margin-based explanation against Breiman's doubts by proving a new generalization error bound that considers exactly the same factors as \cite{Schapire:Freund:Bartlett:Lee1998} but is sharper than \cite{Breiman1999}'s minimum margin bound. By incorporating factors such as average margin and variance, we present a generalization error bound that is heavily related to the whole margin distribution. We also provide margin distribution bounds for generalization error of voting classifiers in finite VC-dimension space.
\end{abstract}

\begin{keyword}
classification \sep boosting \sep ensemble methods \sep margin theory
\end{keyword}

\end{frontmatter}

\section{Introduction}\label{sec:intro}

The \texttt{AdaBoost} algorithm \citep{Freund:Schapire1996a,Freund:Schapire1997}, which aims to construct a ``strong'' classifier by combining some ``weak'' learners (slightly better than random guess), is a representative of ensemble methods \citep{Zhou2012} and has been one of the most influential classification algorithms \citep{Caruana:Niculescu-Mizil2006,Wu:Kumar2009}, and it has exhibited excellent performance both on benchmark datasets and real applications \citep{Bauer:Kohavi1999,Dietterich2000}.

Many studies are devoted to understanding the mysteries behind the success of \texttt{AdaBoost}, among which the margin theory proposed by \cite{Schapire:Freund:Bartlett:Lee1998} has been very influential. For example, \texttt{AdaBoost} often tends to be empirically resistant (but not completely) to overfitting \citep{Quinlan1996,Drucker:Cortes1996,Breiman1998}, i.e., the generalization error of the combined learner keeps decreasing as its size becomes very large and even after the training error has reached zero; it seems violating the Occam's razor \citep{Blumer:Ehrenfeucht:Haussler:Warmuth1987}, i.e., the principle that less complex classifiers should perform better. This remains one of the most famous mysteries of \texttt{AdaBoost}. The margin theory provides the most intuitive and popular explanation to this mystery, that is: \texttt{AdaBoost} tends to improve the margin even after the error on training sample reaches zero.

However, \cite{Breiman1999} raised serious doubt on the margin theory by designing \texttt{arc-gv}, a boosting-style algorithm. This algorithm is able to maximize the \textit{minimum margin}, i.e., the smallest margin over the training data (The formal definition will be given in Eqn.~\ref{eq:minmargin}), but its generalization error is high on empirical datasets, and similar experimental evidence has also been observed in \citep{Grove:Schuurmans1998}. Thus, \cite{Breiman1999} concluded that the margin theory for \texttt{AdaBoost} failed. Breiman's argument was backed up with a minimum margin bound, which is sharper than the generalization bound given by \cite{Schapire:Freund:Bartlett:Lee1998}, and a lot of experiments. \cite{Garg:Roth2003} presented a margin-distribution algorithm based on a data-dependent complexity measure. Later, \cite{Reyzin:Schapire2006} found that there were flaws in the design of experiments: Breiman used CART trees \citep{Breiman:Friedman:Olshen:Stone1984} as base learners and fixed the number of leaves for controlling the complexity of base learners. However, \cite{Reyzin:Schapire2006} found that the trees produced by \texttt{arc-gv} were usually much deeper than those produced by \texttt{AdaBoost}. Generally, for two trees with the same number of leaves, the deeper one is with a larger complexity because more judgements are needed for making a prediction. Therefore, \cite{Reyzin:Schapire2006} concluded that Breiman's observation was biased due to the poor control of model complexity. They repeated the experiments by using decision stumps for base learners, considering that decision stump has exactly two leaves and thus with a fixed complexity, and observed that though \texttt{arc-gv} produced a larger minimum margin, its margin distribution was quite poor. Nowadays, it is well-accepted that the margin distribution is crucial to relate margin to the generalization performance of \texttt{AdaBoost}. To support the margin theory, \cite{Wang:Sugiyama:Yang:Zhou:Feng2011} presented a sharper bound in term of \textit{Emargin} (the formal definition will be given in Theorem~\ref{thm:wang}), which was believed to be relevant to margin distribution.

In this paper, we first present the \textit{$k$th margin} bound and further study its relationship to previous work such as the minimum margin bound and Emargin bound. Then, by using empirical Bernstein bounds, we present a new generalization error bound for voting classifier, which considers exactly the same factors as \cite{Schapire:Freund:Bartlett:Lee1998}, but is sharper than the bounds of \cite{Schapire:Freund:Bartlett:Lee1998} and \cite{Breiman1999}. Therefore, we defend the margin-based explanation against Breiman's doubt. Moreover, we provide a generalization error bound, by incorporating other factors such as average margin and variance, which are heavily relevant to the whole margin distribution. We also give a margin distribution bound for generalization error of voting classifiers in finite VC-dimension space. It is also worth mentioning that our new empirical Bernstein bounds improve the main results of \citep{Maurer:Pontil2009,Audibert:Munos:Szepesvari2009}, with a simpler proof, and we present empirical Bernstein bounds for finite VC-dimension space; these results can be interesting, independently to the main purpose of the paper, to the machine learning community.

The rest of this paper is organized as follows. We begin with some notations and background in Sections~\ref{sec:Notations} and \ref{sec:back}, respectively. Then, we prove the $k$th margin bound and discuss on its relation to previous bounds in Section~\ref{sec:kmargin}. Our main results are presented in Section~\ref{sec:mainresult}, and detailed proofs are provided in Section~\ref{sec:pf}. We conclude in Section \ref{sec:con}.

\section{Notations}\label{sec:Notations}

Let $\mathcal {X}$ and $\mathcal {Y}$ denote an input space and output space, respectively. In this paper, we focus on binary classification problems, i.e., $\mathcal{Y}=\{+1,-1\}$. Denote by $D$ an (unknown) underlying probability distribution over the product space $\mathcal {X}\times\mathcal {Y}$. A training sample of size $m$
\[
S=\{(x_1,y_1), (x_2,y_2), \cdots, (x_m,y_m)\}
\]
is drawn independently and identically (i.i.d) according to the distribution $D$. We use $\Pr_{D}[\cdot]$ to refer as the probability with respect to $D$, and $\Pr_{S}[\cdot]$ to denote the probability with respect to uniform distribution over the sample $S$. Similarly, we use $E_{D}[\cdot]$ and $E_{S} [\cdot]$ to denote the expected values, respectively. For an integer $m>0$, we set $[m]=\{1,2,\cdots,m\}$.

The Bernoulli Kullback-Leibler (or KL) divergence is defined as
\[
KL(q||p)= q\log\frac{q}{p}+(1-q)\log\frac{1-q}{1-p} \text{ for }0\leq p,q\leq 1.
\]
For a fixed $q$, we can easily find that $KL(q||p)$ is a monotone increasing function for $q\leq p<1$, and thus, the inverse of $KL(q||p)$ for the fixed $q$ is given by
\[
KL^{-1}(q;u)=\inf_{w}\left\{w\colon w\geq q\text{ and } KL(q||w)\geq u\right\}.
\]

Let $\mathcal{H}$ be a hypothesis space.  A base learner is a function which maps a distribution over $\mathcal {X}\times\mathcal {Y}$ onto a function $h\colon\mathcal{X} \rightarrow \mathcal{Y}$. In this paper, we only focus on binary base classifiers, i.e., the outputs are in $\{-1, 1\}$. Let $\mathcal{C}(\mathcal{H})$ denote the convex hull of $H$, i.e., a voting classifier $f\in \mathcal{C}(\mathcal{H})$ is of the following form
\[
f=\sum \alpha_ih_i \text{ with }\sum \alpha_i=1 \text{ and }\alpha_i\geq0.
\]
For $N\geq1$,  denote by $\mathcal{C}_N(\mathcal{H})$ the set of unweighted averages over $N$ elements from $\mathcal{H}$, that is
\begin{equation}\label{eq:CN(H)}
\mathcal{C}_N(\mathcal{H})=\Big\{g\colon g=\sum_{j=1}^N \frac{h_j}{N},h_j\in \mathcal{H}\Big\}.
\end{equation}
For voting classifier $f\in \mathcal{C}(\mathcal{H})$, we can associate with a distribution over $\mathcal{H}$ by using the coefficients $\{\alpha_i\}$, denoted by $\mathcal{Q}(f)$. For convenience, $g\in \mathcal{C}_N(\mathcal{H}) \sim\mathcal{Q}(f)$ implies $g=\sum_{j=1}^N {h_j}/{N}$ where $h_j\sim\mathcal{Q}(f)$.

For an example $(x,y)$, the \emph{margin} with respect to the voting classifier $f=\sum \alpha_i h_i(x)$ is defined as $yf(x)$; in other words,
\[
yf(x)=\sum_{i\colon y=h_i(x)}\alpha_i - \sum_{i\colon y\neq h_i(x)}\alpha_i,
\]
which shows the difference between the weights of base learners that classify $(x,y)$ correctly and the weights of base learners that misclassify $(x,y)$. Therefore, margin can be viewed as a measure of the confidence of the classification. Given a sample $S=\{(x_1,y_1), (x_2,y_2), \cdots, (x_m,y_m)\}$, we denote by $\hat{y}_1f(\hat{x}_1)$ the \textit{minimum margin} and $E_S[yf(x)]$ the \textit{average margin}, which are defined respectively as follows:
\begin{equation}\label{eq:minmargin}
\hat{y}_1f(\hat{x}_1)=\min_{i\in[m]}\{y_if(x_i)\} \ \ \text{ and }\ \ E_S[yf(x)]=\sum_{i=1}^m \frac{y_if(x_i)}{m}.
\end{equation}

\section{Background}\label{sec:back}
In the statistics community, great efforts have been devoted to understanding how and why \texttt{AdaBoost} works. \cite{Friedman:Hastie:Tibshirani2000} made an important stride by viewing \texttt{AdaBoost} as a stagewise optimization and relating it to fitting an additive logistic regression model. Various new boosting-style algorithms were developed by performing a gradient decent optimization of some potential loss functions \citep{Mason:Baxter:Bartlett:Frean1999,Ratsch:Onoda:Muller2001,Buhlmann:Yu2003}. Based on this optimization view, some boosting-style algorithms and their variants have been shown to be Bayes's consistent under different settings \citep{Breiman2000tech,Jiang2004,Zhang2004ann,Lugosi:Vayatis2004,Bartlett:Jordan:McAuliffe2006,Bickel:Ritov:Zakai2006,Bartlett:Traskin2007,Mukherjee:Rudin:Schapire2011}, i.e., those studies theoretically ensure that boosting is asymptotically convergent to the Bayes's classifiers. However, such theories can not be used to explain the resistance of \texttt{AdaBoost} to overfitting for small sample problems, and some statistical views have been questioned by \cite{Mease:Wyner2008} with empirical evidences. In this paper, we focus on margin theory.

\begin{algorithm}[!t]\label{alg}
\caption{A unified description of \texttt{AdaBoost} and \texttt{arc-gv}} \textbf{Input}: Sample $S=\{(x_1,y_1),(x_2,y_2),\cdots,(x_m,y_m)\}$ and the number of iterations $T$.\vspace{+2mm}\\
\textbf{Initialization}: $D_1(i)=1/m$.

\begin{algorithmic}
\FOR{$t=1$ to $T$}
\STATE 1. Construct base learner $h_t\colon \mathcal{X}\to\mathcal{Y}$ using the distribution $D_t$.
\STATE 2. Choose $\alpha_t$.
\STATE 3. Update
\[
D_{t+1}(i)={D_t(i)\exp(-\alpha_ty_ih_t(x_i))}/{Z_t},
\]
where $Z_t$ is a normalization factor (such that $D_{t+1}$ is a distribution). \ENDFOR
\end{algorithmic}\vspace{+2mm}
\textbf{Output}: The final classifier $\text{sgn}[f(x)]$, where
 \[
 f(x)=\sum_{t=1}^T \frac{\alpha_t}{\sum_{t=1}^T\alpha_t}h_t(x).
 \]
\end{algorithm}

Algorithm 1 provides a unified description of \texttt{AdaBoost} and \texttt{arc-gv}. The only difference between them lies in the choice of $\alpha_t$. In \texttt{AdaBoost}, $\alpha_t$ is chosen by
\[
\alpha_t=\frac{1}{2}\ln\frac{1+\gamma_t}{1-\gamma_t},
\]
where $\gamma_t= \sum_{i=1}^m D_t(i)y_ih_t(x_i)$ is called the \textit{edge} of $h_t$, which is an affine transformation of the error rate of $h_t(x)$. However, \texttt{Arc-gv} sets $\alpha_t$ in a different way. Denote by $\rho_t$ the minimum margin of the voting classifier of round $t-1$, that is,
\[
\rho_t=\hat{y}_1 f_t(\hat{x}_1)\text{ with }\rho_1=0
\]
where
\[
f_t=\sum_{s=1}^{t-1}\frac{\alpha_s}{\sum_{s=1}^{t-1} \alpha_s}h_s(x).
\]
Then, \texttt{Arc-gv} sets $\alpha_t$ as to be
\[
\alpha_t=\frac{1}{2}\ln\frac{1+\gamma_t}{1-\gamma_t}- \frac{1}{2}\ln\frac{1+\rho_t}{1-\rho_t}.
\]

\cite{Schapire:Freund:Bartlett:Lee1998} proposed the first margin theory for \texttt{AdaBoost} and upper bounded the generalization error as follows:
\begin{theorem}\citep{Schapire:Freund:Bartlett:Lee1998}\label{thm:Scha}
For any $\delta>0$ and $\theta>0$, with probability at least $1-\delta$ over the random choice of sample $S$ with size $m$, every voting classifier $f\in \mathcal{C}(\mathcal{H})$ satisfies the following bound:
\[
\Pr_D[yf(x)<0]\leq \Pr_S[yf(x)\leq\theta]+O\left( \frac{1}{\sqrt{m}} \left(\frac{\ln m\ln|\mathcal{H}|}{\theta^2}+\ln\frac{1}{\delta}\right)^{1/2}\right).
\]
\end{theorem}
\cite{Breiman1999} provided the minimum margin bound for \texttt{arc-gv} by Theorem~\ref{thm:brei} with our notations.
\begin{theorem}\citep{Breiman1999}\label{thm:brei}
If
\[
\theta=\hat{y}_1f(\hat{x}_1)>4\sqrt{\frac{2}{|\mathcal{H}|}}\text{ and } R=\frac{32\ln2|\mathcal{H}|} {m\theta^2}\leq2m,
\]
then, for any $\delta>0$, with probability at least $1-\delta$ over the random choice of sample $S$ with size $m$, every voting classifier $f\in \mathcal{C}(\mathcal{H})$ satisfies the following bound:
\[
\Pr_D[yf(x)<0]\leq R\Big(\ln(2m)+ \ln\frac{1}{R}+1\Big)+ \frac{1}{m}\ln\frac{|\mathcal{H}|}{\delta}.
\]
\end{theorem}
Empirical results show that \texttt{arc-gv} probably generates a larger minimum margin but with higher generalization error, and Breiman's minimum bound is $O({\ln m}/{m})$, sharper than $O(\sqrt{{\ln m}/{m}})$ in Theorem~\ref{thm:Scha}. Thus, Breiman cast serious doubt on margin theory. To support the margin theory, \cite{Wang:Sugiyama:Yang:Zhou:Feng2011} presented a sharper bound in term of \textit{Emargin} by Theorem~\ref{thm:wang}, which was believed to be related to margin distribution. Notice that the factors considered by \cite{Wang:Sugiyama:Yang:Zhou:Feng2011} are different from that considered by \cite{Schapire:Freund:Bartlett:Lee1998} and \cite{Breiman1999}.

\begin{theorem}\citep{Wang:Sugiyama:Yang:Zhou:Feng2011}\label{thm:wang}
If $8<|\mathcal{H}|<\infty$, then for any $\delta>0$, with probability at least $1-\delta$ over the random choice of the training set $S$ of size $m>1$, every voting classifier $f\in \mathcal{C}(\mathcal{H})$ such that
\begin{equation}\label{eq:wangassumption}
q_0=\Pr_S\left[yf(x)\leq\sqrt{8/|\mathcal{H}|}\right]<1
\end{equation}
satisfies the following bound:
\[
\Pr_D[yf(x)<0]\leq \frac{\ln|\mathcal{H}|}{m}+\inf_{q\in\{q_0,q_0+\frac{1}{m},\cdots,1\}} KL^{-1}(q;u[\hat{\theta}(q)]),
\]
where
\[
u[\hat{\theta}(q)]= \frac{1}{m}\Big( \frac{8\ln|\mathcal{H}|}{\hat{\theta}^2(q)} \ln\frac{2m^2}{\ln|\mathcal{H}|} +\ln|\mathcal{H}|+\ln\frac{m}{\delta}\Big)
\]
and $\hat{\theta}(q)=\sup\big\{\theta \in \big(\sqrt{{8}/{|\mathcal{H}|}},1\big]\colon \Pr_S[yf(x)\leq\theta]\leq q\big\}$. Also, the Emargin is given by $\theta^*\in\arg\inf_{q\in\{q_0, q_0+\frac{1}{m},\cdots,1\}} KL^{-1}(q;u[\hat{\theta}(q)])$.
\end{theorem}

Instead of the whole function space, much work developed margin-based data-dependent bounds for generalization error, e.g., empirical cover number \citep{Shawe-Taylor:Williamson1999}, empirical fat-shattering dimension \citep{Antos:Kegl:Linder:Lugosi2002}, Rademacher and Gaussian complexities \citep{Koltchinskii:Panchanko2002,Koltchinskii:Panchanko2005}, etc. Some of these bounds are proven to be sharper than Theorem~\ref{thm:Scha}, but it is hard to show that these bounds are sharper than the bounds of Theorems~\ref{thm:brei} and \ref{thm:wang}, and fail to explain the resistance of \texttt{AdaBoost} to overfitting.

\section{The $k$th Margin Bounds}\label{sec:kmargin}

Given a sample $S$ of size $m$, we define the \textit{$k$th margin} $\hat{y}_kf(\hat{x}_k)$ as the $k$th smallest margin over sample $S$, i.e., the $k$th smallest value in $\{y_if(x_i), i\in[m]\}$. The following theorem shows that the $k$th margin can be used to measure the performance of a voting classifier, whose proof is deferred in Section~\ref{sec:pf1}.
\begin{theorem}\label{thm:kmar}
For any $\delta>0$ and $k\in [m]$, if $\theta=\hat{y}_{k} f(\hat{x}_{k})> \sqrt{{8}/{|\mathcal{H}|}}$, then with probability at least $1-\delta$ over the random choice of sample with size $m$, every voting classifier $f\in \mathcal{C}(\mathcal{H})$ satisfies the following bound:
\begin{equation}\label{eq:kmar:re1}
\Pr_D[yf(x)<0]\leq \frac{\ln|\mathcal{H}|}{m}+KL^{-1}\Big(\frac{k-1}{m};\frac{q}{m}\Big)\,,
\end{equation}
where
\[
q=\frac{8\ln(2|\mathcal{H}|)} {\theta^2}\ln\frac{2m^2} {\ln|\mathcal{H}|}+\ln|\mathcal{H}|+\ln\frac{m}{\delta}.
\]
Particularly, when $k$ is constant with $m>4k$, we have
\begin{equation}\label{eq:kmar:re2}
\Pr_D[yf(x)<0]\leq \frac{\ln|\mathcal{H}|}{m}+\frac{2}{m}\Big(\frac{8\ln(2|\mathcal{H}|)} {\theta^2} \ln\frac{2m^2}{\ln|\mathcal{H}|}+\ln|\mathcal{H}|+\ln\frac{km^{k-1}}{\delta}\Big).
\end{equation}
\end{theorem}

Here, we present the $k$th margin bound to link previous results on margin bounds, and it is interesting to study the relation between Theorem~\ref{thm:kmar} and previous results, especially Theorems~\ref{thm:brei} and \ref{thm:wang}. It is straightforward to get a result similar to Breiman's minimum margin bound in Theorem~\ref{thm:brei}, by setting $k=1$ in Eqn.~\ref{eq:kmar:re2}:
\begin{corollary}\label{coro-brei}
For any $\delta>0$, if $\theta=\hat{y}_{1} f(\hat{x}_{1})>\sqrt{{8}/{|\mathcal{H}|}}$, then with probability at least $1-\delta$ over the random choice of sample $S$ with size $m$, every voting classifier $f\in \mathcal{C}(\mathcal{H})$ satisfies the following bound:
\[
\Pr_D[yf(x)<0]\leq \frac{\ln|\mathcal{H}|}{m} +\frac{2}{m}\Big( \frac{8\ln(2|\mathcal{H}|)}{\theta^2}\ln\frac{2m^2}{\ln|\mathcal{H}|} +\ln\frac{|\mathcal{H}|}{\delta}\Big).
\]
\end{corollary}

Notice that when $k$ is a constant, the bound in Eqn.~\ref{eq:kmar:re2} is $O({\ln m}/{m})$ and the only difference lies in the coefficient. Thus, there is no essential difference to select constant $k$th margin (such as the $2$nd margin, the $3$rd margin, etc.) to measure the confidence of classification for large-size sample.

Based on Theorem~\ref{thm:kmar}, it is not difficult to get a result similar to the Emargin bound in Theorem~\ref{thm:wang} as follows:
\begin{corollary}\label{corol-wang}
For any $\delta>0$, if $\theta_k=\hat{y}_kf(\hat{x}_k)>\sqrt{8/|\mathcal{H}|}$, then with probability at least $1-\delta$ over the random choice of the sample $S$ with size $m$, every voting classifier $f\in \mathcal{C}(\mathcal{H})$ satisfies the following bound:
\[
\Pr_D[yf(x)<0]\leq \frac{\ln|\mathcal{H}|}{m}+ \inf_{k\in[m]}KL^{-1} \Big(\frac{k-1}{m};\frac{q}{m}\Big),
\]
where
\[
q=\frac{8\ln(2|\mathcal{H}|)} {\theta_k^2}\ln\frac{2m^2} {\ln|\mathcal{H}|}+\ln|\mathcal{H}|+\ln\frac{m}{\delta}.
\]
\end{corollary}

From this corollary, we can easily understand that the Emargin bound ought to be tighter than the minimum margin bound because the former takes the infimum over all $k\in[m]$ while the latter only focuses on the minimum margin. Intuitively, the bound of Corollary~\ref{corol-wang} might be sharper than that of Corollary~\ref{coro-brei} if the minimum margin is very small whereas some $k$th margin is very large. We also notice that, as shown by Eqn.~\ref{eq:minmargin}, the minimum margin can also be expressed as taking the infimum over all margin, whereas it is well accepted that the minimum margin bound is a single-margin bound.

\section{Main Results}\label{sec:mainresult}
We begin with the standard deviation bounds as follows:
\begin{theorem}\label{thm:varbound}
For independent random variables $X_1,X_2,\ldots,X_m$ ($m\geq5$) with values in $[0,1]$, and for $\delta\in(0,1)$, we have
\begin{eqnarray}
\Pr\left[\sqrt{E[\hat{V}_m]}<\sqrt{\hat{V}_m}- \sqrt{\frac{\ln1/\delta}{4m}}\right]&\leq&\delta,\label{eq:var_tmp2}\\
\Pr\left[\sqrt{E[\hat{V}_m]}>\sqrt{\hat{V}_m} +\sqrt{\frac{2\ln1/\delta}{m}}\right]&\leq&\delta,\label{eq:var_tmp1}
\end{eqnarray}
where the sample variance $\hat{V}_m=\sum_{i\neq j}{(X_i-X_j)^2}/{2m(m-1)}$.
\end{theorem}

The detailed proof is presented in Section~\ref{sec:pf:thm:varbound}. This theorem improves the results of \citep[Theorem 10]{Maurer:Pontil2009}, especially for Eqn.~\ref{eq:var_tmp2}. Based on this result, we can derive the following empirical Bernstein bounds, with proof deferred to Section~\ref{sec:pf:tool}.
\begin{theorem}\label{thm:tool}
For independent random variables $X_1, X_2, \ldots, X_m$ ($m\geq5$) with values in $[0,1]$, and for $\delta\in(0,1)$, with probability at least $1-\delta$ we have
\begin{eqnarray}
\frac{1}{m}\sum_{i=1}^mE[X_i]-\frac{1}{m}\sum_{i=1}^m X_i &\leq& \sqrt{\frac{2\hat{V}_m\ln (2/\delta)}{m}} + \frac{7\ln(2/\delta)}{3m}, \label{eq:tool2} \\
\frac{1}{m}\sum_{i=1}^mE[X_i]-\frac{1}{m}\sum_{i=1}^m X_i &\geq& -\sqrt{\frac{2\hat{V}_m\ln (2/\delta)}{m}} - \frac{7\ln(2/\delta)}{3m}, \label{eq:tool1}
\end{eqnarray}
where $\hat{V}_m=\sum_{i\neq j}(X_i-X_j)^2/2m(m-1)$.
\end{theorem}
For identical and independent distribution (i.i.d) variables, we have
\begin{corollary}\label{coro:tool}
For i.i.d. random variables $X,X_1, X_2, \ldots, X_m$ ($m\geq5$) with values in $[0,1]$, and for $\delta\in(0,1)$, with probability at least $1-\delta$ we have
\begin{eqnarray*}
E[X]-\frac{1}{m}\sum_{i=1}^m X_i &\leq& \sqrt{\frac{2\hat{V}_m\ln (2/\delta)}{m}} + \frac{7\ln(2/\delta)}{3m}, \\
E[X]-\frac{1}{m}\sum_{i=1}^m X_i &\geq& -\sqrt{\frac{2\hat{V}_m\ln (2/\delta)}{m}} - \frac{7\ln(2/\delta)}{3m},
\end{eqnarray*}
where $\hat{V}_m=\sum_{i\neq j}(X_i-X_j)^2/2m(m-1)$.
\end{corollary}

There are two results \citep{Audibert:Munos:Szepesvari2009,Maurer:Pontil2009} closely related to Theorem~\ref{thm:tool} (or Corollary~\ref{coro:tool}). \cite{Audibert:Munos:Szepesvari2009} presented the first empirical Bernstein bound and applied to analyze multi-armed bandit algorithms. Soon after, \cite{Maurer:Pontil2009} improved the constants and explored the sample variance penalization methods. Comparing with these results, our bounds in Eqns.~\ref{eq:tool2} and \ref{eq:tool1} are with better constants and the technique of proof is simpler.

Based on this Corollary~\ref{coro:tool}, we can derive the following corollary for the finite function space:
\begin{corollary}
Let $S=\{X_1,\ldots,X_m\}$ $(m\geq5)$ be drawn i.i.d. from a distribution $\mathcal{D}$ over $\mathcal{X}$, and let $\mathcal{H}=\{h\colon \mathcal{X}\to[0,1]\}$ be a finite function space. For any $\delta\in(0,1)$, every $h\in\mathcal{H}$ satisfies the following bound with probability at least $1-\delta$:
\[
E_{\mathcal{D}}[h(X)]-\frac{1}{m}\sum_{i=1}^m h(X_i)\leq \sqrt{\frac{2\hat{V}_m(h)\ln (2|\mathcal{H}|/\delta)}{m}} + \frac{7\ln(2|\mathcal{H}|/\delta)}{3m}
\]
where $\hat{V}_m(h)=\sum_{i\neq j}(h(X_i)-h(X_j))^2/2m(m-1)$.
\end{corollary}

Then, we get a new generalization bound for infinite hypothesis space with finite VC-dimension, with proof deferred to Section~\ref{sec:pf:vcdim}.
\begin{theorem}\label{thm:vcdimen}
Let $S=\{X_1,\ldots,X_m\}$ $(m\geq5)$ be drawn i.i.d. from a distribution $\mathcal{D}$ over $\mathcal{X}$, and let $\mathcal{H}=\{h\colon \mathcal{X}\to\{0,1\}\}$ be a hypothesis space with finite VC-dimension $d$. For any $\delta\in(0,1)$, every $h\in\mathcal{H}$ satisfies the following bound with probability at least $1-\delta$:
\[
E_{\mathcal{D}}[h(X)]-\sum_{i=1}^m\frac{h(X_i)}{m} \leq \sqrt{\frac{2\hat{V}_m(h)}{m}\Big(d\ln\frac{2m}{d}+\ln\frac{8}{\delta}\Big)} +\frac{19}{3m}\Big(d\ln\frac{2m}{d}+\ln\frac{8}{\delta}\Big)
\]
where $\hat{V}_m(h)=\sum_{i\neq j}(h(X_i)-h(X_j))^2/2m(m-1)$.
\end{theorem}

\

We now present our first margin bound for AdaBoost as follows:
\begin{theorem}\label{thm:our1}
For any $\delta>0$, with probability at least $1-\delta$ over the random choice of sample $S$ with size $m\geq5$, every voting classifier $f\in \mathcal{C}(\mathcal{H})$ satisfies the following bound:
\[
\Pr_D[yf(x)<0] \leq \frac{2}{m}+\inf_{\theta\in (0,1]} \left[\Pr_S[yf(x)<\theta]+ \frac{7\mu+3\sqrt{3\mu}}{3m}+\sqrt{\frac{3\mu}{m}\Pr_S[yf(x)<\theta]}\right]
\]
where
\[
\mu=\frac{8}{\theta^2}\ln m\ln(2|\mathcal{H}|)+ \ln\frac{2|\mathcal{H}|}{\delta}.
\]
\end{theorem}

This proof is based on the techniques developed by \cite{Schapire:Freund:Bartlett:Lee1998}, and the main difference is that we utilize the empirical Bernstein bound of Eqn.~\ref{eq:tool2} in Theorem~\ref{thm:tool} for the derivation of generalization error. The detailed proof is deferred to Section~\ref{sec:pf2}.

It is noteworthy that Theorem~\ref{thm:our1} shows that the generalization error can be bounded in term of the empirical margin distribution $\Pr_S[yf(x)\leq\theta]$, the training sample size and the hypothesis complexity; in other words, this bound considers exactly the same factors as \cite{Schapire:Freund:Bartlett:Lee1998} in Theorem~\ref{thm:Scha}. However, the following corollary shows that, the bound in Theorem~\ref{thm:our1} is sharper than the bound of \cite{Schapire:Freund:Bartlett:Lee1998} in Theorem~\ref{thm:Scha}, as well as the minimum margin bound of \cite{Breiman1999} in Theorem~\ref{thm:brei}.

\begin{corollary}\label{coro:tighter}
For any $\delta>0$, if the minimum margin $\theta_1=\hat{y}_1f(\hat{x}_1)>0$ and $m\geq5$, then we have
\begin{multline}\label{eq:tight1}
\inf_{\theta\in (0,1]} \left[\Pr_S[yf(x)<\theta]+ \frac{7\mu+3\sqrt{3\mu}}{3m}+ \sqrt{\frac{3\mu}{m}\Pr_S[yf(x)<\theta]}\right]\\
\leq {7\mu_1/3m+\sqrt{3\mu_1}/m},
\end{multline}
where $\mu={8\ln m}\ln(2|\mathcal{H}|)/{\theta^2}+ \ln({2|\mathcal{H}|}/ {\delta})$ and $\mu_1={8\ln m}\ln(2|\mathcal{H}|)/{\theta_1^2}+ \ln({2|\mathcal{H}|}/ {\delta})$; moreover, if
\begin{eqnarray}
&\theta_1=\hat{y}_1f(\hat{x}_1)>4 \sqrt{\frac{2}{|\mathcal{H}|}}\label{eq:t1},&\\
&R=\frac{32\ln2|\mathcal{H}|} {m\theta_1^2}\leq2m\label{eq:t2},&\\
&m\geq \max\Big\{4, \exp\Big(\frac{\theta_1^2}{4\ln(2|\mathcal{H}|)} \ln\frac{|\mathcal{H}|}{\delta}\Big)\Big\}\label{eq:con},&
\end{eqnarray}
then we have
\begin{multline}\label{eq:tight2}
\frac{2}{m}+\inf_{\theta\in (0,1]} \left[\Pr_S[yf(x)<\theta]+ \frac{7\mu+ 3 \sqrt{3\mu}}{3m}+  \sqrt{\frac{3\mu}{m}\Pr_S[yf(x)<\theta]}\right]\\
\leq R\Big(\ln(2m)+ \ln\frac{1}{R}+1\Big)+ \frac{1}{m} \ln\frac{|\mathcal{H}|}{\delta}.
\end{multline}
\end{corollary}

This proof is deferred to Section~\ref{sec:pf4}. From Eqn.~\ref{eq:tight1}, we can see clearly that the bound of Theorem~\ref{thm:our1} is $O(\ln m/m)$, sharper than the bound of \cite{Schapire:Freund:Bartlett:Lee1998} $O(\sqrt{\ln m/m})$ in Theorem~\ref{thm:Scha}. In fact, we could also guarantee that bound of Theorem~\ref{thm:our1} is $O(\ln m/m)$ even under weaker assumption that $\hat{y}_kf(\hat{x}_k)>0$ for some $k\leq O(\ln m)$.

It is also noteworthy Eqns.~\ref{eq:t1} and \ref{eq:t2} are the conditions of Theorem~\ref{thm:brei}, and the term $\exp \Big( \frac{\theta_1^2} {4\ln(2|\mathcal{H}|)}\ln\frac{|\mathcal{H}|}{\delta}\Big)\leq (\frac{e}{\delta})^\frac{1}{4}$ in Eqn.~\ref{eq:con}, which is small for many real applications, e.g., it is less than $13$ even if $\delta=0.0001$. Eqn.~\ref{eq:tight2} shows that the bound of Theorem~\ref{thm:our1} is sharper than Breiman's minimum margin bound of Theorem~\ref{thm:brei}.

\cite{Breiman1999} doubted the margin theory because of two recognitions: i) the minimum margin bound of \cite{Breiman1999} is sharper than the margin distribution bound of \cite{Schapire:Freund:Bartlett:Lee1998}, and therefore, the minimum margin is more essential than margin distribution to characterize the generalization performance; ii) \texttt{arc-gv} maximizes the minimum margin, but demonstrates worse performance than \texttt{AdaBoost} empirically. However, our result shows that the margin distribution bound in Theorem~\ref{thm:Scha} can be greatly improved such that it is even sharper than the minimum margin bound, and therefore, it is natural that \texttt{AdaBoost} outperforms \texttt{arc-gv} empirically on some datasets; in a word, our results provide a complete answer to Breiman's doubt on margin theory.

The Emargin bounds of \citet{Wang:Sugiyama:Yang:Zhou:Feng2011} are also proven to be sharper than those of \citet{Schapire:Freund:Bartlett:Lee1998} and \citet{Breiman1999}. The main difference between Theorem~\ref{thm:our1} and the Emargin bounds lies in the consideration of different factors for margin theory, e.g., Theorem~\ref{thm:our1} considers exactly the same factors as \cite{Schapire:Freund:Bartlett:Lee1998}, whereas \citet{Wang:Sugiyama:Yang:Zhou:Feng2011} considered the Emargin as the key factor. Moreover, Theorem~\ref{thm:our1} is advantageous in that its margin interval is wider than that of Emargin\footnote{This observation owes to a reviewer}. Note that it is not easy to directly compare Theorem~\ref{thm:our1} and the Emargin bounds because it is difficult to get a closed-form for the $D^{-1}(p||q)$ term contained in the Emargin bounds, whereas Theorem~\ref{thm:our1} is relatively easier to estimate.

\

It is well-accepted that the margin distribution is crucial to relate margin to the generalization performance of \texttt{AdaBoost}, whereas it is unclear how to measure the ``goodness'' of a margin distribution. The first-order and second-order statistics, i.e., the average margin and variance, are natural and intuitive measures. Indeed, \citet{Reyzin:Schapire2006} has recommended to take the average margin for a characterization for the margin distribution. However, there is no theory, to the best of our knowledge, to support that a larger average margin or a smaller variance implies a smaller generalization error. The following theorem fills the gap for such theory:
\begin{theorem}\label{thm:our2}
For any $\delta>0$, with probability at least $1-\delta$ over the random choice of sample $S$ with size $m\geq5$, every voting classifier $f\in \mathcal{C}(\mathcal{H})$ satisfies the following bound:
\begin{multline*}
\Pr_D[yf(x)<0] \leq \frac{1}{m^{50}}+\inf_{\theta\in (0,1]}\left[\Pr_S[yf(x)<\theta]+m^{-2/(1-E^2_{S}[yf(x)]+\theta/9)}\right.\\
\left. + \frac{3\sqrt{\mu}}{m^{3/2}}+ \frac{7\mu}{3m} + \sqrt{\frac{3\mu}{m}\hat{\mathcal{I}}(\theta)}\right]
\end{multline*}
where
\begin{eqnarray*}
&&\mu={144}\ln m\ln(2|\mathcal{H}|)/{\theta^2}+\ln({2|\mathcal{H}|}/\delta),\\
&&\hat{\mathcal{I}}(\theta)=\Pr_S[yf(x)<\theta] \Pr_S[yf(x)\geq2\theta/3].
\end{eqnarray*}
\end{theorem}

The detailed proof is deferred to Section~\ref{sec:pf3}. It is easy to find in almost all boosting experiments that the average margin $E_S[yf(x)]$ is positive. Thus, the bound of Theorem~\ref{thm:our2} can be sharper for larger average margin. The statistics $\hat{\mathcal{I}}(\cdot)$ reflects the margin variance in some sense, and the term including $\hat{\mathcal{I}}(\cdot)$ can be small or even vanished except for a small interval when the variance is small. This new generalization error bound depends not only on the sample size and the complexity of base classifiers, but also on the average margin, variance, and empirical margin distribution; this implying that, completely explaining AdaBoost's resistance to overfitting is more difficult than what has been expected and disclosed by previous theoretical results.

\begin{figure}[!t]
\begin{center}
\includegraphics[width=6.5in]{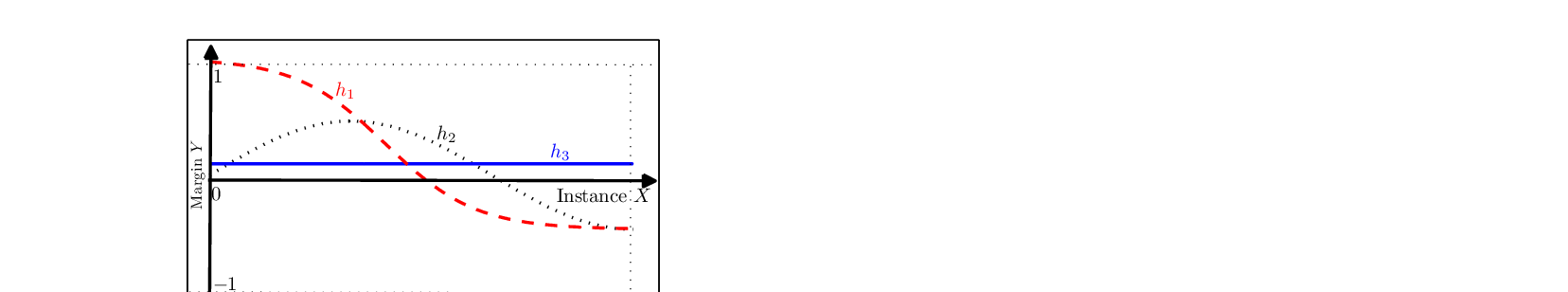}
\caption{Each curve represents a voting classifier. The $X$-axis and $Y$-axis denote example and margin, respectively, and uniform distribution is assumed on the example space. The voting classifiers $h_1$, $h_2$ and $h_3$ have the same average margin but with different generalization error rates: ${1}/{2}$, ${1}/{3}$ and $0$.}\label{fig1}
\end{center}
\end{figure}

Theorem~\ref{thm:our2} also provides a theoretical support to the suggestion of \cite{Reyzin:Schapire2006}; that is, the average margin can be used to measure the performance. It is noteworthy that, however, merely considering the average margin is insufficient to bound the generalization error tightly, as shown by the simple example in Figure~\ref{fig1}. Indeed, as this theorem discloses, ``average'' and ``variance'' are two important statistics to capture a distribution, and it is reasonable that both the average margin and margin variance are considered.

We have the following corollary with proof presented in Section~\ref{sec:pf:corowholemargintighter}.
\begin{corollary}\label{coro:wholemaringtighter}
If the minimum margin $\theta_1=\hat{y}_1f(\hat{x}_1)>0$, then, for any $\delta>0$, with probability at least $1-\delta$ over the random choice of sample $S$ with size $m\geq5$, every voting classifier $f\in \mathcal{C}(\mathcal{H})$ satisfies the following bound:
\begin{multline*}
\frac{1}{m^{50}}+\inf_{\theta\in (0,1]}\left[\Pr_S[yf(x)<\theta] +m^{-2/(1-E^2_{S}[yf(x)]+\theta/9)}\right.\\
\left.+ \frac{3 \sqrt{\mu}}{m^{3/2}}+\frac{7\mu}{3m}+ \sqrt{\frac{3\mu}{m}\hat{\mathcal{I}}(\theta)}\right] \leq \frac{1}{m^{50}}+\frac{1}{m^{2}}+ \frac{3\sqrt{\mu_1}}{m^{3/2}}+ \frac{7\mu_1}{3m}
\end{multline*}
where $\mu_1={144}\ln m\ln(2|\mathcal{H}|)/{\theta_1^2}+ \ln({2|\mathcal{H}|}/\delta)$, $\mu$ and $\hat{\mathcal{I}}(\theta)$ are given in Theorem~\ref{thm:our2}.
\end{corollary}

This corollary shows that the bounds of Theorem~\ref{thm:our2} are $O(\ln m/m)$, comparable to the Emargin bounds \citep{Wang:Sugiyama:Yang:Zhou:Feng2011} and the bounds of Theorem~\ref{thm:our1}, but with different constants. The main difference lies in the consideration of different factors, as we have considered the average margin and variance, that are better for the characterization of margin distribution. It is noteworthy that the best bounds for \texttt{AdaBoost} and \texttt{arc-gv} are both $O(\ln m/m)$ whereas \texttt{AdaBoost} outperforms \texttt{arc-gv} empirically because \texttt{AdaBoost} tends to improve the margin distribution; this provides an example showing that it is very important to consider factors that are heavily relevant to the whole distribution. We also notice that a recent study in \citep{Shen:Li2010} provides empirical evidence to support our theoretical result. Indeed, designing new Boosting algorithms that maximize average margin but minimize variance simultaneously is an interesting direction, and \citep{Shivaswamy:Jebara2011} may shed some light.

\

Finally, we generalize our main margin bounds to the case when the space of base classifiers has finite VC-dimension. The detailed proofs are presented in Section~\ref{sec:pf:thm:vcdim}.
\begin{theorem}\label{thm:our1vcdim}
If the base classifiers space $\mathcal{H}$ has finite VC-dimension $d$, then for any $\delta>0$, with probability at least $1-\delta$ over the random choice of sample $S$ with size $m\geq5$, every voting classifier $f\in \mathcal{C}(\mathcal{H})$ satisfies the following bound:
\[
\Pr_D[yf(x)<0] \leq \frac{2}{m}+\inf_{\theta\in (0,1]} \left[\Pr_S[yf(x)<\theta]+ \frac{19\mu+3\sqrt{3\mu}}{3m}+\sqrt{\frac{3\mu}{m}\Pr_S[yf(x)<\theta]}\right]
\]
where $\mu=\frac{8\ln m}{\theta^2}\big(\ln2+d\ln(2em/d)\big) + \ln\big(\frac{8}{\delta}(1+\frac{8\ln m}{\theta^2})\big)$.
\end{theorem}

\begin{theorem}\label{thm:our2vcdim}
If the base classifiers space $\mathcal{H}$ has finite VC-dimension $d$, then for any $\delta>0$, with probability at least $1-\delta$ over the random choice of sample $S$ with size $m\geq5$, every voting classifier $f\in \mathcal{C}(\mathcal{H})$ satisfies the following bound:
\begin{multline*}
\Pr_D[yf(x)<0] \leq \frac{1}{m^{50}}+\inf_{\theta\in (0,1]}\left[\Pr_S[yf(x)<\theta] +m^{-2/(1-E^2_{S}[yf(x)]+\theta/9)}\right.\\
\left. + \frac{3\sqrt{\mu}}{m^{3/2}}+ \frac{19\mu}{3m}+ \sqrt{\frac{3\mu}{m}\hat{\mathcal{I}}(\theta)}\right]
\end{multline*}
where
\begin{eqnarray*}
&&\mu={144}\big(\ln2+d\ln(2em/d)\big)\ln m/{\theta^2}+ \ln\big((8+576\ln m/\theta^2 )/\delta\big),\\
&&\hat{\mathcal{I}}(\theta)=\Pr_S[yf(x)<\theta] \Pr_S[yf(x)\geq2\theta/3].
\end{eqnarray*}
\end{theorem}

\section{Proofs}\label{sec:pf}
In this section, we provide the detailed proofs for the main theorems and corollaries. First, we present a series of useful lemmas as follows:

\begin{lemma}[Chernoff bound \citep{Chernoff1952}]\label{lem:Chern} Let $X, X_1, X_2, \ldots, X_m$ be $m+1$ i.i.d random variables with $X\in[0,1]$. Then, for any $\epsilon>0$, we have
\begin{eqnarray*}
&&\Pr\left[\frac{1}{m}\sum_{i=1}^m X_i\geq E[X]+\epsilon\right] \leq\exp\left(-\frac{m\epsilon^2}{2}\right),\\
&&\Pr\left[\frac{1}{m}\sum_{i=1}^mX_i\leq E[X]-\epsilon\right]\leq \exp\left(-\frac{m\epsilon^2}{2}\right).
\end{eqnarray*}
\end{lemma}

\begin{lemma}[Relative entropy Chernoff bound \citep{Hoeffding1963}]\label{lem:entropy}
For $0<\epsilon<1$, we have
\[
\sum_{i=0}^{k-1}{m\choose i} \epsilon^i(1-\epsilon)^{m-i} \leq \exp\left(-m KL\left(\frac{k-1}{m}\Big|\Big|\epsilon\right)\right).
\]
\end{lemma}

\begin{lemma}[Bennett's inequalities \citep{McDiarmid1998}]\label{lem:ben} For independent random variables $X, X_1, X_2, \ldots, X_m$ with $X_i\in[0,1]$, and for any $\delta>0$, the followings hold with probability at least $1-\delta$
\begin{eqnarray}
\frac{1}{m}\sum_{i=1}^m E[X_i]-\frac{1}{m}\sum_{i=1}^mX_i&\leq& \sqrt{\frac{2V(X)\ln1/\delta}{m}}+\frac{\ln1/\delta}{3m},\label{eq:bernineq1}\\
\frac{1}{m}\sum_{i=1}^m E[X_i]-\frac{1}{m}\sum_{i=1}^mX_i&\geq& -\sqrt{\frac{2V(X)\ln1/\delta}{m}}-\frac{\ln1/\delta}{3m}, \label{eq:bernineq2}
\end{eqnarray}
where $V(X)$ denotes the variance $\sum_{i=1}^mE[(X_i-E[X_i])^2]/m$.
\end{lemma}

\subsection{Proof of Theorem~\ref{thm:kmar}}\label{sec:pf1}

We begin with a lemma as follows:
\begin{lemma}\label{lem-temp0}
For $f\in\mathcal{C}(\mathcal{H})$, let $g\in \mathcal{C}_N(\mathcal{H})$ be drawn i.i.d according to distribution $\mathcal{Q}(f)$. If $\hat{y}_{k} f(\hat{x}_{k})\geq \theta$ and $\hat{y}_{k} g(\hat{x}_{k})\leq \alpha$ with $\theta>\alpha$, then there is an example $(x_i,y_i)$ in $S$ such that $y_{i} f(x_{i})\geq \theta$ and $y_{i} g(x_{i})\leq \alpha$.
\end{lemma}
\pf There exists a bijection between $\{y_jf(x_j)\colon j\in[m]\}$ and $\{y_jg(x_j)\colon j\in [m]\}$ according to the original position in $S$. Suppose $\hat{y}_{k} f(\hat{x}_{k})$ corresponds to $\hat{y}_lg(\hat{x}_l)$ for some $l$. If $l\leq k$ then the example $(\hat{x}_k,\hat{y}_k)$ of $\hat{y}_{k} f(\hat{x}_{k})$ is desired; otherwise, except for $(\hat{x}_k,\hat{y}_k)$ of $\hat{y}_{k} f(\hat{x}_{k})$ in $S$, there are at least $m-k$ elements larger than or equal to $\theta$ in $\{y_jf(x_j)\colon j\in[m]\setminus\{k\} \}$ but at most $m-k-1$ elements larger than $\alpha$ in $\{y_jg(x_j)\colon j\in[m]\setminus\{l\}\}$. This completes the proof from the bijection.\qed\vspace{0.15in}

\noindent\textit{Proof of Theorem~\ref{thm:kmar}} For finite $\mathcal{H}$, we denote by $\mathcal{A}=\{{i}/{|\mathcal{H}|}\colon i\in[|\mathcal{H}|]\}$. For every $f\in\mathcal{C}(\mathcal{H})$, we can construct a $g\in \mathcal{C}_N(\mathcal{H})$ by choosing $N$ elements i.i.d according to distribution $\mathcal{Q}(f)$, and thus $E_{g\sim\mathcal{Q}(f)}[g]=f$. For $\alpha>0$, the Chernoff's bound in Lemma~\ref{lem:Chern} gives
\begin{eqnarray}
\Pr_D[yf(x)<0]&=&\Pr_{D,\mathcal{Q}(f)}[yf(x)<0,yg(x)\geq\alpha]+ \Pr_{D,\mathcal{Q}(f)}[yf(x)<0,yg(x)<\alpha]\nonumber\\
&\leq&\exp(-{N\alpha^2}/{2})+\Pr_{D,\mathcal{Q}(f)}[yg(x)<\alpha].\label{eq:thm1:temp0}
\end{eqnarray}
For any $\epsilon_N>0$, we consider the following probability:
\begin{eqnarray}
&&\Pr_{S\sim D^m}\left[\Pr_D[yg(x)<\alpha]> I[\hat{y}_{k} g(\hat{x}_{k}) \leq \alpha]+\epsilon_N \right]\nonumber \\
&&\leq  \Pr_{S\sim D^m} \left[\hat{y}_{k} g(\hat{x}_{k})>\alpha \left|\Pr_D[yg(x)<\alpha] >\epsilon_N\right.\right]\nonumber \\
&&\leq \sum_{i=0}^{k-1}{m\choose i} \epsilon_N^i(1-\epsilon_N)^{m-i} \label{eq:thm1:temp1}
\end{eqnarray}
where $\hat{y}_{k} g(\hat{x}_{k})$ denotes the $k$th margin with respect to $g$. For any $k$, Eqn.~\ref{eq:thm1:temp1} can be bounded by $\exp\big(-m KL\big(\frac{k-1}{m} \big|\big|\epsilon_N\big)\big)$ from Lemma~\ref{lem:entropy}; for constant $k$ with $m > 4k$, we have
\begin{eqnarray*}
\sum_{i=0}^{k-1}{m\choose i} \epsilon_N^i(1-\epsilon_N)^{m-i} &\leq&k(1-\epsilon_N)^{m/2} {m\choose k-1}\\
&\leq&km^{k-1}(1-\epsilon_N)^{m/2}\leq km^{k-1}e^{-\epsilon_Nm/2}.
\end{eqnarray*}
By using the union bound and $|\mathcal{C}_N(\mathcal{H})|\leq |\mathcal{H}|^{N}$, we have, for any $k\in[m]$,
\begin{eqnarray*}
&&\Pr_{S\sim {D}^m,g\sim \mathcal{Q}(f)}\left[\exists g\in \mathcal{C}_N(\mathcal{H}), \exists \alpha \in \mathcal{A},\Pr_D[yg(x)<\alpha]> I[\hat{y}_{k} g(\hat{x}_{k})\leq \alpha]+\epsilon_N\right]\\
&&\quad\quad \leq |\mathcal{H}|^{N+1} \exp\left(-m KL\Big(\frac{k-1}{m}\big|\big|\epsilon_N\Big)\right).
\end{eqnarray*}
Setting $\delta_N=|\mathcal{H}|^{N+1} \exp\big(-m KL\big(\frac{k-1}{m} \big|\big| \epsilon_N\big) \big)$ gives
\[
\epsilon_N=KL^{-1}\big(\frac{k-1}{m};\frac{1}{m} \ln\frac{|\mathcal{H}|^{N+1}}{\delta_N}\big).
\]
Thus, with probability at least $1-\delta_N$ over sample $S$, for all $f\in \mathcal{C}(\mathcal{H})$ and all $\alpha\in \mathcal{A}$, we have
\begin{equation}\label{eq:thm1:temp2}
\Pr_D[yg(x)<\alpha]\leq I[\hat{y}_{k} g(\hat{x}_{k}) \leq \alpha]
+KL^{-1}\left(\frac{k-1}{m};\frac{1}{m} \ln\frac{|\mathcal{H}|^{N+1}}{\delta_N}\right).
\end{equation}
Similarly, for constant $k$, with probability at least $1-\delta_N$ over sample $S$, it holds that
\begin{equation}\label{eq:thm1:temp3}
\Pr_D[yg(x)<\alpha]\leq I[\hat{y}_{k} g(\hat{x}_{k})\leq\alpha]
+\frac{2}{m}\ln\frac{km^{k-1}|\mathcal{H}|^{N+1}}{\delta_N}.
\end{equation}
From $E_{g\sim \mathcal{Q}(f)}[I[\hat{y}_{k} g(\hat{x}_{k})\leq \alpha]] =\Pr_{ g \sim \mathcal{Q}(f)}[\hat{y}_{k} g(\hat{x}_{k})\leq \alpha]$, we have, for any $\theta>\alpha$,
\begin{multline}\label{eq:thm1:temp5}
\Pr_{g\sim \mathcal{Q}(f)}[\hat{y}_{k} g(\hat{x}_{k})\leq \alpha]\leq I[\hat{y}_{k} f(\hat{x}_{k})<\theta]\\
+\Pr_{g\sim \mathcal{Q}(f)}[\hat{y}_{k} f(\hat{x}_{k})\geq \theta, \hat{y}_{k}g(\hat{x}_{k})\leq\alpha].
\end{multline}
Notice that the example $(\hat{x}_k,\hat{y}_k)$ in $\{\hat{y}_if(\hat{x}_i)\}$ may be different from example $(\hat{x}_k,\hat{y}_k)$ in $\{\hat{y}_ig(\hat{x}_i)\}$; therefore, we can not bound the last term on the right-hand side of Eqn.~\ref{eq:thm1:temp5} as done in \citep{Wang:Sugiyama:Yang:Zhou:Feng2011}, whereas it can be bounded by using Lemma \ref{lem-temp0}
\begin{equation}\label{eq:thm1:temp6}
\Pr_{g\sim \mathcal{Q}(f)}[\exists (x_i,y_i)\in S\colon y_{i} f(x_{i})\geq \theta, y_{i} g(x_{i})\leq \alpha]\leq m\exp(-N(\theta-\alpha)^2/2).
\end{equation}
Combining Eqns.~\ref{eq:thm1:temp0}, \ref{eq:thm1:temp2}, \ref{eq:thm1:temp5} and \ref{eq:thm1:temp6}, we have that with probability at least $1-\delta_N$ over the sample $S$, for all $f\in \mathcal{C}(\mathcal{H})$, all $\theta>\alpha$, all $k\in[m]$ but fixed $N$:
\begin{multline}\label{eq:thm1:temp7}
\Pr_D[yf(x)<0]\leq I[\hat{y}_{k} f(\hat{x}_{k})\leq\theta] +m\exp(-{N(\theta-\alpha)^2}/{2}) +\exp(-N\alpha^2/2)\\
+KL^{-1}\left(\frac{k-1}{m};\frac{1}{m} \ln\frac{|\mathcal{H}|^{N+1}m}{\delta_N}\right).
\end{multline}
To obtain the probability of failure for any $N$ at most $\delta$, we select $\delta_N=\delta/2^N$. Setting $\alpha=\frac{\theta}{2}-\frac{\eta}{|\mathcal{H}|}\in \mathcal{A}$ and $N=\lceil\frac{8}{\theta^2}\ln\frac{2m^2}{\ln|\mathcal{H}|}\rceil$  with $0\leq\eta<1$, we have
\[
\exp(-N\alpha^2/2)+m\exp(-N(\theta-\alpha)^2/2)\leq 2m\exp(-N\theta^2/8)\leq \ln|\mathcal{H}|/m
\]
from the fact $2m>\exp(N/(2|\mathcal{H}|))$ for $\theta>\sqrt{8/|\mathcal{H}|}$. Finally we obtain
\[
\Pr[yf(x)<0]\leq I[\hat{y}_{k} f(\hat{x}_{k}) < \theta]+ \frac{\ln|\mathcal{H}|}{m} +KL^{-1}\left(\frac{k-1}{m}||\frac{q}{m}\right)
\]
where $q=\frac{8\ln(2|\mathcal{H}|)} {\theta^2}\ln\frac{2m^2} {\ln|\mathcal{H}|}+\ln|\mathcal{H}|+\ln\frac{m}{\delta}$. This completes the proof of Eqn.~\ref{eq:kmar:re1}. In a similar manner, we have
\begin{multline*}
\Pr[yf(x)<0]\leq I[\hat{y}_{k} f(\hat{x}_{k}) < \theta]+{\ln|\mathcal{H}|}/{m}\\
+\frac{2}{m}\left(\frac{8\ln(2|\mathcal{H}|)} {\theta^2}\ln\frac{2m^2}{\ln|\mathcal{H}|}+\ln|\mathcal{H}|+\ln\frac{km^{k-1}}{\delta}\right),
\end{multline*}
for constant $k<m/4$. This completes the proof of Eqn.~\ref{eq:kmar:re2} as desired.\qed

\subsection{Proof of Theorem~\ref{thm:varbound}}\label{sec:pf:thm:varbound}
For notational simplicity, we denote by $\bar{X}=(X_1,X_2,\ldots,X_m)$ a vector of $m$ i.i.d. random variables, and further set
\[
\bar{X}^{k,Y}=(X_1,\ldots, X_{k-1}, Y, X_{k+1},\ldots,X_m),
 \]
i.e., the vector with the the $k$th variable $X_k$ in $\bar{X}$ replaced by variable $Y$. We first introduce some lemmas as follows:

\begin{lemma}[McDiarmid Formula \citep{McDiarmid89}]\label{McDiarmid} Suppose that $\bar{X}=(X_1,X_2,\ldots,X_m)$ is a vector of $m$ i.i.d. random variables taking values in a set $\mathcal{A}$. If $|F(\bar{X})- F(\bar{X}^{k,Y})|\leq c_k$ for $k\in [m]$ and $Y\in\mathcal{A}$, then the following holds for any $t>0$,
\[
\Pr\left[F(\bar{X})-E[F(\bar{X})]\geq t\right]\leq \exp\left(\frac{-2t^2}{\sum_{k=1}^mc_k^2}\right).
\]
\end{lemma}

\begin{lemma}[Theorem 13 \citep{Maurer2006}]\label{lem:mau}
Let $\bar{X}=(X_1,X_2,\ldots,X_m)$ be a vector of $m$ independent random variables tanking values in a set $\mathcal{A}$. If $F\colon\mathcal{A}^m\to \mathbb{R}$ satisfies that
\[
F(\bar{X})-\inf_{Y\in\mathcal{A}}F(\bar{X}^{k,Y}) \leq 1 \text{ and } \sum_{k=1}^m \left(F(\bar{X})- \inf_{Y\in\mathcal{A}}F(\bar{X}^{k,Y})\right)^2 \leq F(\bar{X}),
\]
then for any $t>0$, we have
\[
\Pr[E[F(\bar{X})]-F(\bar{X})>t]\leq \exp({-t^2}/{2 E[F(\bar{X})]}).
\]
\end{lemma}

\begin{lemma}\label{lem:tmp1}
For two i.i.d random variables $X$ and $Y$, we have
\[
E[(X-Y)^2]=2E[(X-E[X])^2]=2V(X).
\]
\end{lemma}
\pf This lemma follows from the obvious fact $E[(X-Y)^2]= E(X^2+Y^2-2XY)= 2E[X^2]-2E^2[X]= 2E[(X-E[X])^2]$.\qed\vspace{0.15in}

\noindent\textit{Proof of Theorem~\ref{thm:varbound}} We will utilize Lemmas~\ref{McDiarmid} and \ref{lem:mau} to prove Eqns.~\ref{eq:var_tmp2} and \ref{eq:var_tmp1}, respectively. For Eqn.~\ref{eq:var_tmp2}, we first observe that, for any $k\in[m]$,
\[
\left|\sqrt{\hat{V}_m(\bar{X})}-\sqrt{\hat{V}_m(\bar{X}^{k,Y})}\right| =\left|\frac{{\hat{V}_m(\bar{X})}-{\hat{V}_m(\bar{X}^{k,Y})}} {\sqrt{\hat{V}_m(\bar{X})}+\sqrt{\hat{V}_m(\bar{X}^{k,Y})}}\right|\leq \frac{1}{\sqrt{2}m},
\]
where we use $\hat{V}_m(\bar{X}),\hat{V}_m(\bar{X}^{k,Y})\leq 1/2$ from $X_i\in[0,1]$. By using the Jenson's inequality, we have $E\big[\sqrt{\hat{V}_m(\bar{X})}\big]\leq \sqrt{E[\hat{V}_m(\bar{X})]}$ and thus,
\begin{eqnarray*}
&&\Pr\left[\sqrt{E[\hat{V}_m(\bar{X})]}<\sqrt{\hat{V}_m(\bar{X})}-\epsilon\right]\\
&&\leq \Pr\left[ E\left[\sqrt{\hat{V}_m(\bar{X})}\right]<\sqrt{\hat{V}_m(\bar{X})}- \epsilon\right]\\
&&\leq \exp(-4m\epsilon^2).
\end{eqnarray*}
where the last inequality holds by applying McDiarmid formula in Lemma~\ref{McDiarmid} to $\sqrt{\hat{V}_m}$. Therefore, we complete the proof of Eqn.~\ref{eq:var_tmp2} by setting $\delta=\exp(-4m\epsilon^2)$. \vspace{0.08in}

To prove Eqn.~\ref{eq:var_tmp1}, we set $\xi_m(\bar{X})=m\hat{V}_m(\bar{X})$. For $X_i\in [0,1]$ and $\xi_m(\bar{X}^{k,Y})$, it is easy to obtain the optimal solution by simple calculation
\[
Y^*={\arg\inf}_{Y\in[0,1]}[\xi_m(\bar{X}^{k,Y})]=\sum\nolimits_{i\neq k}\frac{X_i}{m-1},
\]
which yields that
\begin{eqnarray*}
\xi_m(\bar{X})-\inf_{Y\in[0,1]}[\xi_m(\bar{X}^{k,Y})]&=&\frac{1}{m-1}\sum_{i\neq k}(X_i-X_k)^2-(Y^*-X_i)^2\\
&=&\Big(X_k-\sum_{i\neq k}\frac{X_i}{m-1}\Big)^2.
\end{eqnarray*}
For $X_i\in [0,1]$, it is obvious that
\[
\xi_m(\bar{X})- \inf_{Y\in[0,1]} [\xi_m(\bar{X}^{k,Y})]\leq 1,
\]
and from Lemma~\ref{lem:tmp1}, we have
\[
\frac{1}{m} \sum_{k=1}^m \Big(X_k-\sum_{i=1}^m\frac{X_i}{m}\Big)^2 \leq \frac{1}{2m^2}\sum_{i,k}(X_i-X_k)^2 = \frac{1}{2m^2}\sum_{i\neq k}(X_i-X_k)^2,
\]
which yields that, for $m\geq5$,
\[
\sum_{k=1}^m(\xi_m(\bar{X})-\inf_{Y\in[0,1]}[\xi_m(\bar{X}^{k,Y})])^2 \leq \frac{m^3}{4(m-1)^4}\sum_{i\neq k}(X_i-X_k)^2\leq \xi_m(\bar{X}).
\]
Therefore, for any $t>0$, the following holds by using Lemma~\ref{lem:mau} to $\xi_m(\bar{X})$,
\begin{eqnarray*}
\Pr[E[\hat{V}_m(\bar{X})]-\hat{V}_m(\bar{X})>t]&=& \Pr[E[\xi_m(\bar{X})]-\xi_m(\bar{X})>mt] \\
&\leq& \exp\left({-mt^2}/{2E[\hat{V}_m(\bar{X})]}\right).
\end{eqnarray*}
Setting $\delta=\exp({-mt^2}/{2E[\hat{V}_m(\bar{X})]})$ gives
\[
\Pr\left[E[\hat{V}_m(\bar{X})] - \hat{V}_m(\bar{X}) > \sqrt{{2E[\hat{V}_m(\bar{X})]\ln(1/\delta)}/m}\right]\leq \delta
\]
which completes the proof of Eqn.~\ref{eq:var_tmp1} by using the square-root's inequality and $\sqrt{a+b}\leq \sqrt{a}+\sqrt{b}$ for $a,b\geq0$. \qed \vspace{0.15in}

\subsection{Proof of Theorem~\ref{thm:tool}}\label{sec:pf:tool}
For independent random variables $\bar{X}=(X_1,X_2,\ldots, X_m)$, we set $\hat{V}_m(\bar{X})=\sum_{i\neq j} (X_i-X_j)^2/2m(m-1)$, and observe that
\begin{eqnarray*}
\lefteqn{E[\hat{V}_m(\bar{X})]=\frac{1}{2m(m-1)}\sum_{i\neq j}E[(X_i-X_j)^2]}\\
&=&\frac{1}{2m(m-1)}\sum_{i\neq j}\left(E[(X_i-E[X_i])^2]+ E[(X_j-E[X_j])^2+(E[X_i]-E[X_j])^2 \right)\\
&\geq&\frac{1}{m}\sum_{i}E(X_i-E[X_i])^2=V,
\end{eqnarray*}
where we denote by $V=\sum_{i}E(X_i-E[X_i])^2/m$ and the second equality holds from $(a+b+c)^2=a^2+b^2+c^2+2ab+2ac+2bc$. For any $\delta>0$, the following holds with probability at least $1-\delta$ from Eqn.~\ref{eq:bernineq1},
\[
\frac{1}{m}\sum_{i=1}^m (E[X_i]-X_i) \leq \sqrt{\frac{2V\ln1/\delta}{m}}+ \frac{\ln1/\delta}{3m}\leq\sqrt{\frac{2E[\hat{V}_m(\bar{X})]\ln1/\delta}{m}}+ \frac{\ln1/\delta}{3m}
\]
which completes the proof of Eqn.~\ref{eq:tool2} by combining with Eqn.~\ref{eq:var_tmp1} in a union bound and simple calculations. Similar proof could be made for Eqn.~\ref{eq:tool1}. \qed

\subsection{Proof of Theorem~\ref{thm:vcdimen}}\label{sec:pf:vcdim}
We will use classical double sample method \citep{Devroye:Gyorfi:Lugosi1996,vapnik98} to prove Theorem~\ref{thm:vcdimen}. Let $\mathscr{A}$ be a subsets of space $\mathcal{Z}$, and we define
\[
s(\mathscr{A},m)=\max\{|\{A\cap S\colon A\in\mathscr{A}\}|\colon S\subseteq \mathcal{Z} \text{ and }|S|=m\}.
\]
We first introduce a useful lemma as follows:
\begin{lemma}\label{lem:vcdim}
For space $\mathscr{A}$ of subsets of $\mathcal{Z}$, and for sample $S=(z_1,z_2,\ldots,z_m)$ drawn i.i.d. from distribution $\mathcal{D}$ over $\mathcal{Z}$, we have, for $t>\ln4$
\[
\Pr_{S\sim\mathcal{D}^m}\left[\exists A\in \mathscr{A}\colon \Pr_\mathcal{D}[A] >\Pr_S[A] +\sqrt{\frac{2t}{m}\hat{V}_S(A)} +\frac{19t}{3m} \right]\leq 8s(\mathscr{A},2m)e^{-t}
\]
where $\Pr_\mathcal{D}[A]=\Pr_{z\sim \mathcal{D}}[z\in A]$, $\Pr_S[A]=\Pr_{z\sim S}[z\in A]$ and $\hat{V}_S(A)=\sum_{i\neq j}(I[z_i\in A]-I[z_j\in A])^2/2m(m-1)$.
\end{lemma}
\pf We begin with another sample $\hat{S}=(\hat{z}_1,\hat{z}_2,\ldots,\hat{z}_m)$ drawn identically and independently from distribution $\mathcal{D}$, and denote by
\[
\Psi_S(A)=\Pr\nolimits_{{S}}[A]+ \sqrt{2\hat{V}_{{S}}(A)t/m}+7t/3m.
\]
From Corollary~\ref{coro:tool}, we have $\Pr_{\hat{S}\sim\mathcal{D}^m} [\Pr_\mathcal{D}[A] \leq \Psi_{\hat{S}}(A)]\geq1/2$ for $h\in\mathcal{H}$ and $t>\ln4$. This follows for any $\epsilon>0$
\begin{eqnarray*}
\lefteqn{\Pr_{S\sim\mathcal{D}^m}\left[\exists A\in\mathscr{A}\colon \Pr_\mathcal{D}[A]> \Psi_S(A)+\epsilon\right]}\\
&=&E_{S\sim\mathcal{D}^m}\sup_{A\in\mathscr{A}}I\left[\Pr_\mathcal{D}[A]> \Psi_S(A)+\epsilon\right]\\
&\leq& 2E_{S\sim\mathcal{D}^m}\sup_{A\in\mathscr{A}}I\left[\Pr_\mathcal{D}[A]> \Psi_S(A)+\epsilon\right] E_{\hat{S}\sim\mathcal{D}^m}I\left[\Pr_\mathcal{D}[A]\leq \Psi_{\hat{S}}(A)\right]\\
&\leq&2\Pr_{S\sim\mathcal{D}^m,\hat{S}\sim\mathcal{D}^m}\left[\exists A\in\mathscr{A}\colon \Psi_{\hat{S}}(A)>\Psi_S(A)+\epsilon \right].
\end{eqnarray*}
Now, we introduce the sign random variable vector $\sigma=(\sigma_1, \sigma_2, \ldots, \sigma_m)$ with probability $\Pr[\sigma_i=1]=\Pr[\sigma_i=-1]=1/2$ for $i\in[m]$, and denote by $S^\sigma=(z^\sigma_i)_{i=1}^m$ and $\hat{S}^\sigma=(\hat{z}^\sigma_i)_{i=1}^m$
\[
z^\sigma_i=z_i, \hat{z}^\sigma_i=\hat{z}_i \text{ if }\sigma=1; \text{ otherwise, } z^\sigma_i=\hat{z}_i,\hat{z}_i^{\sigma_i}=z_i.
\]
Given $S$ and $S'$, $z^\sigma_i$ ($i\in[m]$) are not identically distributed but independent. Conditioned on $S$ and $S'$, we have
\begin{eqnarray*}
\lefteqn{\Pr_\sigma\left[\exists A\in\mathscr{A}\colon \Psi_{\hat{S}^\sigma}(A)>\Psi_{S^\sigma}(A)+\epsilon |S,S'\right]} \\
&\leq&s(\mathscr{A},2m)\sup_{A\in\mathscr{A}}\Pr_\sigma \left[\Psi_{{\hat{S}^\sigma}}(A)>\Psi_{S^\sigma}(A)+\epsilon|S,S'\right]\\
&=&s(\mathscr{A},2m)\Pr_\sigma \left[\Psi_{{\hat{S}^\sigma}}(A^*)>\Psi_{S^\sigma}(A^*)+\epsilon|S,S'\right]\\
&=&s(\mathscr{A},2m)\Pr_\sigma \left[\Psi_{{\hat{S}^\sigma}}(A^*)>\Pr_\sigma[A^*]+\epsilon/2|S,S'\right]\\
&&+s(\mathscr{A},2m)\Pr_\sigma \left[\Pr_\sigma[A^*]>\Psi_{S^\sigma}(A^*)+\epsilon/2|S,S'\right],
\end{eqnarray*}
where we denote by $A^*\in\arg\sup_{A\in\mathscr{A}}\Pr_\sigma \left[\Psi_{\hat{S}^\sigma}(A)>\Psi_{S^\sigma}(A)+\epsilon|S,S'\right]$ and $\Pr_\sigma[A^*]=E_\sigma[\Pr_{S^\sigma}[A^*]|S,\hat{S}]= E_\sigma[\Pr_{\hat{S}^\sigma}[A^*]|S,\hat{S}]$. Further, we denote by
\[
V_\sigma(A^*)=E_{S^\sigma}[\hat{V}_{S^\sigma}(A^*)|S,\hat{S}] =E_{\hat{S}^\sigma}[\hat{V}_{\hat{S}^\sigma}(A^*)|S,\hat{S}].
\]
Thus, we have
\begin{eqnarray*}
\lefteqn{\Pr_\sigma \left[\Psi_{\hat{S}^\sigma}(A^*)>\Pr_\sigma[A^*]+\epsilon/2|S,S'\right]}\\
&=&\Pr_\sigma \left[\Pr_{\hat{S}^\sigma}[A^*]+ \sqrt{2\hat{V}_{\hat{S}^\sigma}(A^*)t/m}+7t/3m>\Pr_\sigma[A^*]+\epsilon/2|S,S'\right]\\
&\leq& \Pr_\sigma \left[\Pr_{\hat{S}^\sigma}[A^*]+ \sqrt{2V_\sigma(A^*)t/m}+7t/3m>\Pr_\sigma[A^*]|S,S'\right]\\
&&+\Pr_\sigma \left[\sqrt{2\hat{V}_{\hat{S}^\sigma}(A^*)t/m} >\sqrt{2V_\sigma(A^*)t/m}+\epsilon/2|S,S'\right].
\end{eqnarray*}
The first term in the above can be bounded by $e^{-t}$ from Bennett's inequality (Lemma~\ref{lem:ben}), and the second term can be bound by $e^{-t}$ by setting $\epsilon=4t/m$ and using Theorem~\ref{thm:varbound}. Similarly, we can prove
\[
\Pr_\sigma \left[\Pr_\sigma[A^*]>\Psi_{S^\sigma}(A^*)+\epsilon/2|S,S'\right]\leq 2e^{-t}
\]
by setting $\epsilon=4t/m$. This complete the proof as desired.\qed\vspace{0.1in}

\noindent\textit{Proof of Theorem~\ref{thm:vcdimen}:} Let
\[
\mathscr{A}=\{A(h)\colon h\in\mathcal{H}\}
\]
where $A(h)=\{(X,h(X)\in\mathcal{X}\times\{-1,+1\})\}$. For space $\mathcal{H}$ with finite VC-dimension $d$, Sauer's lemma \citep{Sauer1972} gives
\[
s(\mathscr{A},2m)\leq (2em/d)^d.
\]
Combining with Lemma~\ref{lem:vcdim}, we have, for $t\geq\ln4$
\[
\Pr_{S\sim\mathcal{D}^m}\left[\exists h\in \mathcal{H}\colon E_{\mathcal{D}}[h(X)]>\sum_{i=1}^m\frac{h(X_i)}{m} +\sqrt{\frac{2t\hat{V}_S(h)}{m}} +\frac{19t}{3m} \right]\leq 8\Big(\frac{2m}{d}\Big)^de^{-t}.
\]
Setting $\delta=8({2m}/{d})^de^{-t}$, we have
\[
t=d\ln(2m/d)+ln(8/\delta)\geq\ln4 \text{ for }\delta\in(0,1),
\]
which complete the proof.\qed

\subsection{Proof of Theorem~\ref{thm:our1}}\label{sec:pf2}
Similarly to the proof of Theorem~\ref{thm:kmar}, we have
\begin{equation}\label{eq:tt1}
\Pr_D[yf(x)<0]\leq \exp(-{N\alpha^2}/{2})+\Pr_{D,\mathcal{Q}(f)}[yg(x)<\alpha],
\end{equation}
for any given $\alpha>0$, $f\in\mathcal{C}(\mathcal{H})$ and $g\in \mathcal{C}_N (\mathcal{H})$ drawn i.i.d according to $\mathcal{Q}(f)$. Recall that $|\mathcal{C}_N (\mathcal{H})|\leq |\mathcal{H}|^{N}$. Therefore, for any $\delta_N>0$, combining union bound with Eqn.~\ref{eq:tool2} in Theorem~\ref{coro:tool} guarantees that the following holds with probability at least $1-\delta_N$ over sample $S$, for any $g\in \mathcal{C}_N(\mathcal{H})$ and $\alpha\in\mathcal{A}$,
\begin{multline}\label{eq:tt0}
\Pr_D[yg(x)<\alpha]\leq \Pr_S[yg(x)<\alpha]\\
+\sqrt{\frac{2}{m}\hat{V}_m\ln\big(\frac{2}{\delta_N}|\mathcal{H}|^{N+1}\big)}+ \frac{7}{3m}\ln(\frac{2}{\delta_N}|\mathcal{H}|^{N+1}),
\end{multline}
where
\[
\hat{V}_m=\sum_{i\neq j}\frac{(I[y_ig(x_i)<\alpha]- I[y_jg(x_j)<\alpha])^2}{2m(m-1)}.
\]
Furthermore, we have
\[
\sum_{i\neq j}\left(I[y_ig(x_i)<\alpha]- I[y_jg(x_j)<\alpha]\right)^2=2m^2 \Pr_S[yg(x)<\alpha]\Pr_S[yg(x)\geq\alpha],
\]
which yields that
\begin{equation}\label{eq:tt2}
\hat{V}_m=\frac{m}{m-1}\Pr_S[yg(x)<\alpha]\Pr_S[yg(x)\geq\alpha] \leq\frac{3}{2}\Pr_S[yg(x)<\alpha],
\end{equation}
for $m\geq 5$. By using Lemma~\ref{lem:Chern} again, the following holds for any $\theta_1>0$,
\begin{equation}\label{eq:tt4}
\Pr_S[yg(x)<\alpha]\leq \exp(-{N\theta_1^2}/{2})+\Pr_S[yf(x)<\alpha+\theta_1].
\end{equation}
Setting $\theta_1=\alpha=\theta/2$ and combining Eqns.~\ref{eq:tt1}, \ref{eq:tt0}, \ref{eq:tt2} and \ref{eq:tt4}, we have
\begin{eqnarray*}
\Pr_D[yf(x)<0]&\leq& \Pr_S[yf(x)<\theta]+2\exp(-N\theta^2/8) \\
&&+\frac{7\mu}{3m}+\sqrt{\frac{3\mu}{m}\left(\Pr_S[yf(x)<\theta]+ \exp\left(-\frac{N\theta^2}{8}\right)\right)},
\end{eqnarray*}
where $\mu=\ln(2|\mathcal{H}|^{N+1}/\delta_N)$. By utilizing the fact $\sqrt{a+b} \leq \sqrt{a}+ \sqrt{b}$ for $a\geq 0$  and $b\geq0$, we further have
\begin{eqnarray*}
&&\sqrt{\frac{3\mu}{m}\left(\Pr_S[yf(x)<\theta]+\exp\left(-\frac{N\theta^2}{8} \right)\right)}\\
&&\leq \sqrt{\frac{3\mu}{m}\Pr_S[yf(x)<\theta]} +\sqrt{\frac{3\mu}{m}\exp\left(-\frac{N\theta^2}{8}\right)}.
\end{eqnarray*}
Finally, we set $\delta_N=\delta/2^{N}$ so that the probability of failure for any $N$ will be no more than $\delta$. This theorem follows by setting $N=\lceil8\ln m/\theta^2\rceil$. \qed

\subsection{Proof of Corollary~\ref{coro:tighter}}\label{sec:pf4}
If the minimum margin $\theta_1=\hat{y}_1f(\hat{x}_1)>0$, then we have  $\Pr_S[yf(x)<\theta_1]=0$ and further get
\begin{eqnarray}
&&\inf_{\theta\in (0,1]} \left[\Pr_S[yf(x)<\theta]+ \frac{7\mu+3\sqrt{3\mu}}{3m}+ \sqrt{\frac{3\mu}{m}\Pr_S[yf(x)<\theta]}\right] \nonumber\\
&&\leq\Pr_S[yf(x)<\theta_1]+ \frac{7\mu_1+3\sqrt{3\mu_1}}{3m}+ \sqrt{\frac{3\mu_1}{m}\Pr_S[yf(x)<\theta_1]}\nonumber\\
&&=\frac{7\mu_1+3\sqrt{3\mu_1}}{3m},\label{eq:t3}
\end{eqnarray}
where $\mu_1={8\ln m}\ln(2|\mathcal{H}|)/{\theta_1^2}+ \ln({2|\mathcal{H}|}/ {\delta})$. This gives the proof of Eqn.~\ref{eq:tight1}. If $m\geq5$, then we have
\[
\mu_1\geq\frac{8}{\theta_1^2}\ln{m} \ln(2|\mathcal{H}|)\geq8 \text{ leading to }{\sqrt{3\mu_1}} \leq {2\mu_1}/3.
\]
Therefore, the following holds by combining Eqn.~\ref{eq:t3} and the above facts,
\begin{eqnarray*}
&&\frac{2}{m}+\inf_{\theta\in (0,1]} \left[\Pr_S[yf(x)<\theta]+ \frac{7\mu+3\sqrt{3\mu}}{3m}+ \sqrt{\frac{3\mu}{m}\Pr_S[yf(x)<\theta]}\right]\\
&&\leq \frac{2}{m}+\frac{7\mu_1+3\sqrt{2\mu_1}}{3m} \leq\frac{2}{m}+\frac{3\mu_1}{m}=\frac{2}{m}+\frac{24\ln m}{m\theta_1^2} \ln(2|\mathcal{H}|)+\frac{3}{m}\ln\frac{2|\mathcal{H}|}{\delta}\\
&&\leq\frac{8}{m}+\frac{24\ln m}{m\theta_1^2}\ln(2|\mathcal{H}|)+ \frac{3}{m}\ln\frac{|\mathcal{H}|}{\delta}
\leq R\Big(\ln(2m)+ \ln\frac{1}{R}+1\Big)+ \frac{1}{m}\ln\frac{|\mathcal{H}|}{\delta}
\end{eqnarray*}
where the last inequality holds from the conditions of Eqn.~\ref{eq:con} and ${8}/{m}<R$. This completes the proof of Eqn.~\ref{eq:tight2}.\qed

\subsection{Proof of Theorem~\ref{thm:our2}}\label{sec:pf3}
Our proof is based on a new Bernstein-type bound as follows:
\begin{lemma}\label{lem:t2}
For $f\in\mathcal{C}(\mathcal{H})$ and $g\in \mathcal{C}_N(\mathcal{H})$ drawn  i.i.d according to distribution $\mathcal{Q}(f)$, we have
\[
\Pr_{S,g\sim \mathcal{Q}(f)}\left[yg(x)-yf(x)\geq t\right]\leq\exp\left(\frac{-Nt^2}{2-2E^2_{S}[yf(x)]+4t/3}\right).
\]
\end{lemma}
\pf For $\lambda>0$, we utilize the Markov's inequality to have
\begin{eqnarray*}
&&\Pr_{S,g\sim \mathcal{Q}(f)}[yg(x)-yf(x)\geq t]\\
&&=\Pr_{S,g\sim \mathcal{Q}(f)}[(yg(x)-yf(x))N\lambda/2\geq N\lambda t/2]\\
&&\leq\exp\left(-\frac{\lambda Nt}{2}\right) E_{S,g\sim \mathcal{Q}(f)} \left[\exp\left(\frac{\lambda}{2}\sum_{j=1}^N yh_j(x)-yf(x)\right)\right] \\
&&=\exp(-{\lambda Nt}/{2})\prod_{j=1}^N E_{S,h_j\sim \mathcal{Q}(f)}[\exp( \lambda(yh_j(x)-yf(x))/2)],
\end{eqnarray*}
where the last inequality holds from the independence of $h_j$. Notice that
$|yh_j(x)-yf(x)|\leq 2$ from $\mathcal{H}\subseteq\{h\colon \mathcal{X}\to\{-1,+1\}\}$. By using Taylor's expansion, we further get
\begin{eqnarray*}
&&E_{S,h_j\sim \mathcal{Q}(f)}[\exp(\lambda(yh_j(x)-yf(x))/2)]  \\
&&\leq 1+E_{S,h_j\sim \mathcal{Q}(f)}[(yh_j(x)-yf(x))^2](e^\lambda-1-\lambda)/4  \\
&&=1+E_{S}[1-(yf(x))^2](e^\lambda-1-\lambda)/4\\
&&\leq \exp\left((1-E^2_{S}[yf(x)])(e^\lambda-1-\lambda)/4\right),
\end{eqnarray*}
where the last inequality holds from Jensen's inequality and $1+x\leq e^{x}$. Therefore, it holds that
\begin{eqnarray*}
&&\Pr_{S,g\sim \mathcal{Q}(f)}\left[yg(x)-yf(x)\geq t\right]\\
&&\leq \exp\left(N(e^\lambda-1-\lambda)(1-E^2_{S}[yf(x)])/4-\lambda Nt/2\right).
\end{eqnarray*}
If $0<\lambda<3$, then we could use Taylor's expansion again to have
\[
e^{\lambda}-\lambda-1=\sum_{i=2}^\infty\frac{\lambda^i}{i!}\leq \frac{\lambda^2}{2}\sum_{i=0}^\infty\frac{\lambda^m}{3^m} = \frac{\lambda^2}{2(1-\lambda/3)}.
\]
Now by picking $\lambda={t}/(1/2-E^2_{S}[yf(x)]/2+ t/3)$, we have
\[
-\frac{\lambda t}{2}+\frac{\lambda^2(1-E^2_{S}[yf(x)])}{8(1-\lambda/3)} \leq \frac{-t^2}{2-2E^2_{S}[yf(x)]+4t/3},
\]
which completes the proof as desired.\qed\vspace{0.15in}

\noindent\textit{Proof of Theorem~\ref{thm:our2}} This proof is rather similar to the proof of Theorem~\ref{thm:our1}, and we just give main steps. For any $\alpha>0$ and $\delta_N>0$, the following holds with probability at least $1-\delta_N$ over sample $S_m$ ($m\geq5$),
\begin{multline*}
\Pr_D[yf(x)<0]\leq \Pr_S[yg(x)<\alpha]+ \exp(-{N\alpha^2}/{2})\\
+\sqrt{\frac{3\hat{V}^*_m\ln(\frac{2}{\delta_N}|\mathcal{H}|^{N+1})}{m}}+ \frac{7}{3m}\ln(\frac{2}{\delta_N}|\mathcal{H}|^{N+1}),
\end{multline*}
where $\hat{V}^*_m=\Pr_S[yg(x)<\alpha]\Pr_S[yg(x)\geq\alpha]$. For any $\theta_1>0$, we use Lemma~\ref{lem:Chern} to obtain
\begin{multline*}
\hat{V}^*_m=\Pr_S[yg(x)<\alpha]\Pr_S[yg(x)\geq\alpha]\leq 3\exp(-N\theta_1^2/2)\\
+\Pr_S[yf(x)<\alpha+\theta_1]\Pr_S[yf(x)>\alpha-\theta_1].
\end{multline*}
From Lemma~\ref{lem:t2}, it holds that
\[
\Pr_S[yg(x)<\alpha]\leq \Pr_S[yf(x)<\alpha+\theta_1]+\exp\Big({\frac{-N\theta_1^2} {2-2E^2_{S}[yf(x)]+4\theta_1/3}}\Big).
\]
Let $\theta_1=\theta/6$, $\alpha=5\theta/6$, and set $\delta_N=\delta/2^{N}$ so that the probability of failure for any $N$ will be no more than $\delta$. We complete the proof by setting $N=\lceil144\ln m/\theta^2\rceil$ and simple calculation. \qed

\subsection{Proof of Corollary~\ref{coro:wholemaringtighter}}\label{sec:pf:corowholemargintighter}
If the minimum margin $\theta_1=\hat{y}_1f(\hat{x}_1)>0$, then we have  $\Pr_S[yf(x)<\theta_1]=0$ and $\hat{\mathcal{I}}(\theta_1)=\Pr_S[yf(x)<\theta_1] \Pr_S[yf(x)\geq2\theta_1/3]=0$. Further, we have
\begin{eqnarray*}
&&\inf_{\theta\in (0,1]}\left[\Pr_S[yf(x)<\theta]+ \frac{\sqrt{6\mu}}{m^{3/2}}+ \frac{7\mu}{3m}+\sqrt{\frac{3\mu}{m}\hat{\mathcal{I}}(\theta)}+m^{-2/(1-E^2_{S}[yf(x)+\theta/9)}\right]\\
&&\leq \frac{\sqrt{6\mu_1}}{m^{3/2}}+ \frac{7\mu_1}{3m}+m^{-2/ (1-E^2_{S}[yf(x)]+\theta_1/9)}\\
&&\leq \frac{\sqrt{6\mu_1}}{m^{3/2}}+ \frac{7\mu_1}{3m}+\frac{1}{m^2}
\end{eqnarray*}
where $\mu_1={144\ln m}\ln(2|\mathcal{H}|)/{\theta_1^2}+ \ln({2|\mathcal{H}|}/ {\delta})$. This completes the proof.\qed

\subsection{Proof of Theorems~\ref{thm:our1vcdim} and \ref{thm:our2vcdim}}\label{sec:pf:thm:vcdim}
For finite VC-dimension space $\mathcal{H}$, we denote by $\mathcal{A}=\{i/N\colon i\in [N]\}$. Similarly to the proof of Theorem~\ref{thm:kmar}, we have
\begin{equation}\label{eq:tttt1}
\Pr_D[yf(x)<0]\leq \exp(-{N\alpha^2}/{2})+\Pr_{D,\mathcal{Q}(f)}[yg(x)<\alpha],
\end{equation}
for $\alpha\in\mathcal{A}$, $f\in\mathcal{C}(\mathcal{H})$ and $g\in \mathcal{C}_N (\mathcal{H})$ chosen i.i.d according to $\mathcal{Q}(f)$. Define
\[
\mathscr{A}=\{\{(x,y)\in\mathcal{X}\times\{+1,-1\}\colon yg(x)<\alpha\}\colon g\in \mathcal{C}_N (\mathcal{H}), \alpha\in\mathcal{A}\},
\]
and by using Sauer's lemma \citep{Sauer1972}, we have
\begin{equation}\label{eq:tttt2}
s(\mathscr{A},m)\leq(N+1)(em/d)^{Nd}
\end{equation}
for $m>d$. By setting $4s(\mathscr{A},2m)e^{-t}=\delta_N>0$ in Lemma ~\ref{lem:vcdim}, the following holds with probability at least $1-\delta_N$ over sample $S$, for any $g\in \mathcal{C}_N(\mathcal{H})$ and $\alpha\in\mathcal{A}$,
\begin{multline}\label{eq:tttt3}
\Pr_D[yg(x)<\alpha]\leq \Pr_S[yg(x)<\alpha]\\
+\sqrt{\frac{3}{m}\hat{V}_m\ln\Big(\frac{8s(\mathscr{A},2m)}{\delta_N}\Big)}+ \frac{19}{3m}\ln\Big(\frac{8s(\mathscr{A},2m)}{\delta_N}\Big),
\end{multline}
where $\hat{V}^*_m=\Pr_S[yg(x)<\alpha]\Pr_S[yg(x)\geq\alpha]$.\vspace{0.1in}

To prove Theorem~\ref{thm:our1vcdim}, we proceed as the proof of Theorem~\ref{thm:our1}. Setting $\alpha=\theta/2$, we have
\begin{eqnarray*}
\Pr_D[yf(x)<0]&\leq& \Pr_S[yf(x)<\theta]+2\exp(-N\theta^2/8) \\
&&+\frac{19\mu}{3m}+\sqrt{\frac{3\mu}{m}\left(\Pr_S[yf(x)<\theta]+ \exp\left(-\frac{N\theta^2}{8}\right)\right)},
\end{eqnarray*}
where $\mu=\ln(8s(\mathscr{A},2m)/\delta_N)$. This completes the proof by using  $\sqrt{a+b} \leq \sqrt{a}+ \sqrt{b}$ and setting $\delta_N=\delta/2^{N}$ and $N=\lceil8\ln m/\theta^2\rceil$.\vspace{0.1in}

To prove Theorem~\ref{thm:our2vcdim}, we proceed as the proof of Theorem~\ref{thm:our2}. Setting $\alpha=5\theta/6$, we have
\begin{multline*}
\Pr_D[yf(x)<0]\leq \Pr_S[yf(x)<\theta]+\exp(-25N\theta^2/72)+{19\mu}/{3m} \\
+\exp\Big({\frac{-N\theta^2/36} {2-2E^2_{S}[yf(x)]+ 2\theta/9}}\Big)+ \sqrt{\frac{3\mu}{m}\left(\hat{\mathcal{I}}(\theta)+ 3\exp\left(-\frac{N\theta^2}{72}\right)\right)},
\end{multline*}
where $\mu=\ln(8s(\mathscr{A},2m)/\delta_N)$ and $\hat{\mathcal{I}}(\theta)=\Pr_S[yf(x)<\theta] \Pr_S[yf(x)\geq2\theta/3]$. This completes the proof by using  $\sqrt{a+b} \leq \sqrt{a}+ \sqrt{b}$ and setting $\delta_N=\delta/2^{N}$ and $N=\lceil144\ln m/\theta^2\rceil$. \qed

\section{Conclusion}\label{sec:con}

The margin theory provides one of the most intuitive and popular theoretical explanations to \texttt{AdaBoost}. It is well-accepted that the margin distribution is crucial for characterizing the performance of \texttt{AdaBoost}, and it is desirable to theoretically establish generalization bounds based on margin distribution.

In this paper, we first present the \textit{$k$th margin bound} and further study on its relationship to previous work such as the minimum margin bound and Emargin bound. Then, we improve the empirical Bernstein bound with different skills. As our main results, we prove a new generalization bound which considers exactly the same factors as \cite{Schapire:Freund:Bartlett:Lee1998} but is sharper than the bounds of \cite{Schapire:Freund:Bartlett:Lee1998} and \cite{Breiman1999}, and thus provide a complete answer to Breiman's doubt on the margin theory. By incorporating other factors such as average margin and variance, we present another generalization error bound which is heavily related to the whole margin distribution. In addition, we provide margin bounds for generalization error of voting classifiers in finite VC-dimension space. An interesting future issue is to develop new algorithms based on our theory.

\section*{Acknowledgements}
We want to thank the editor and reviewers for helpful comments and suggestions. This work was supported by the National Fundamental Research Program of China (2010CB327903), the National Science Foundation of China (61073097, 61021062), the Jiangsu Province Graduate Students Innovative Research Project (CXZZ11\_0046) and the Nanjing University PhD Students Promoting Program (201301A07).

\bibliographystyle{elsarticle-harv}
\bibliography{reference}
\end{document}